\RequirePackage{import} 
\documentclass[10pt]{article}

\usepackage{times}
\usepackage[utf8]{inputenc}
\usepackage[T1]{fontenc}
\usepackage{graphicx}

\usepackage{multicol,multirow,array}
\usepackage{booktabs}  
\usepackage[skip=5pt]{caption}  
\usepackage{graphicx}  
\usepackage{svg}  
\usepackage{subcaption}  
\usepackage{array}  
\usepackage[shortlabels]{enumitem} 

\let\mathexp=\exp 
\usepackage{gb4e}
\let\gbexp=\exp 
\DeclareRobustCommand{\exp}{\ifmmode\mathexp\else\expandafter\gbexp\fi}
\noautomath

\usepackage{siunitx}
\usepackage{hyperref}
\usepackage{cleveref}
\usepackage{coria-taln2025}
\usepackage[french]{babel} 

\bibpunct{(}{)}{~;~}{a}{,}{,} 

\newcommand{\PreserveBackslash}[1]{\let\temp=\\#1\let\\=\temp}
\newcolumntype{C}[1]{>{\PreserveBackslash\centering}p{#1}}
\newcolumntype{R}[1]{>{\PreserveBackslash\raggedleft}p{#1}}
\newcolumntype{L}[1]{>{\PreserveBackslash\raggedright}p{#1}}

\newcommand{\emin}{E\textsubscript{cog}}

\title{Annotation et modélisation des émotions dans un corpus textuel~: une approche évaluative}

\author{Jonas Noblet\\
  {\small
    (1) \textsc{lidilem}, Université Grenoble Alpes, Bâtiment Stendhal, CS40700, 38058 Grenoble cedex 9, France \\
    (2) Société Ixiade, 11 Rue Aimé Berey, 38000 Grenoble, France \\
    \texttt{
      jonas.noblet@univ-grenoble-alpes.fr \\ 
}}}

\begin{document}
\maketitle

\resume{
    %
    L'émotion est un phénomène capital dans le fonctionnement de l'être humain en société. Elle reste pourtant un sujet encore largement ouvert, notamment dans ses manifestations textuelles.  La présente communication examine un corpus industriel manuellement annoté selon une approche évaluative de l'émotion. Cette conception théorique aujourd’hui peu exploitée propose une perspective différente, en complément des approches traditionnelles. Partant du constat que les annotations que nous avons collectées présentent un fort désaccord, nous avons émis l’hypothèse que celles-ci suivent néanmoins des tendances statistiques stables. Par le biais de modèles de langue entraînés sur ces annotations, nous montrons qu’il est possible de modéliser le processus d’étiquetage, et que la variabilité est guidée par des caractéristiques linguistiques sous-jacentes. Réciproquement, nos résultats indiquent que les modèles de langue semblent en mesure de distinguer les situations émotionnelles sur la base des critères évaluatifs.
}

\abstract{Annotation and modeling of emotions in a textual corpus: an evaluative approach}{
    Emotion is a crucial phenomenon in the functioning of human beings in society. However, it remains a widely open subject, particularly in its textual manifestations. This paper examines an industrial corpus manually annotated following an evaluative approach to emotion. This theoretical framework, which is currently underutilized, offers a different perspective that complements traditional approaches.
    Noting that the annotations we collected exhibit significant disagreement, we hypothesized that they nonetheless follow stable statistical trends.
    Using language models trained on these annotations, we demonstrate that it is possible to model the labeling process and that variability is driven by underlying linguistic features.
    Conversely, our results indicate that language models seem capable of distinguishing emotional situations based on evaluative criteria.
}

\motsClefs
  {Annotation, Émotion, Accord inter-juges, Modèles de langue}
  {Annotation, Emotion, Inter-rater agreement, Language models}

\section{Introduction}

Comme le montre \citet{Damasio1995DescartesError}, l'émotion joue un rôle central dans la cognition humaine~: elle est un composant essentiel au fonctionnement de l'être humain en société.
L'émotion interagit avec l'attention, la mémoire, le raisonnement et la prise de décision (voir \citet{Lemaire2021EmotCog}, \citet{Houwer2010EmotCogRev}, \citet{Fox2018BigRev2} pour des revues). En même temps que se sont développées les sciences cognitives à partir de la deuxième moitié du \textsc{xx}\textsuperscript{e}~siècle, l'étude de l'émotion a elle aussi crû en importance, recueillant plus de financements et générant plus de publications \citep{Dukes2021RiseAffectivism}.

La présente communication est le fruit d'une collaboration entre le laboratoire \textsc{lidilem} et la société Ixiade, dans le cadre de la thèse \textsc{cifre} menée par \citet{Noblet2026These}. Les recherches dont nous présentons les résultats s'appuient sur un corpus issu des projets\footnote{Nous réservons le terme \textit{projet} pour désigner spécifiquement les études de test de concepts --~bien ou service~-- innovants conduites par Ixiade ; chaque projet est associé à un concept innovant, porté par un client donné \citep{Noblet2025EminosaNoise}.} industriels menés dans le cadre de l'activité économique de l'entreprise. Ces projets ont pour visée d'établir l'acceptabilité\footnote{La notion d’acceptabilité désigne l’usage potentiel d’une innovation telle que l’individu ciblé se la représente, avant les premières manipulations \citep{Terrade2009SocAccept}.} de concepts innovants en sondant l'opinion à leur sujet.
Malgré la nature spécifique de ce corpus, nous proposons une réflexion en mesure de contribuer plus largement à la compréhension des émotions et de leur traitement dans le texte.

La suite de l'article se divise en trois parties. La première introduit des éléments de contexte pour notre étude qui orientent son propos. La seconde présente les données -- corpus et annotation -- sur lesquelles se base la présente contribution. Enfin, la dernière détaille les dernières analyses que nous avons conduites, notamment à l'aide d'un modèle de langue fine-tuné.

\section{Éléments de contexte}

\subsection*{L'émotion, un champ disciplinaire encore débattu}

Depuis son développement jusqu'à aujourd'hui, une théorie majeure domine le champ d'étude des émotions~: la théorie des émotions de base (\textit{basic emotion theory}) développée par Paul Ekman \citep*{Ekman1984OriginCite, Ekman1992BasicEmots, Ekman2011BasicEmotsDef}. Celle-ci, conçue dans le prolongement des travaux de \citet{Darwin1890EmotOrigin} et du mouvement béhavioriste, postule l'existence d'un ensemble d'émotions élémentaires, biologiquement évoluées. Selon cette conception, les émotions prennent la forme de catégories clairement délimitées et universelles, communes à tous les êtres humains. Une version qui s'est beaucoup diffusée propose six émotions de base~: la colère, le dégoût, la peur, la joie, la tristesse et la surprise \citep{Ekman1987UniversalsFace}.

Peu après son apparition, cette théorie fait l'objet de plusieurs critiques. Ce sont à la fois les méthodes expérimentales et les prémices --~l'émotion comme phénomène universel divisé en catégories distinctes~-- qui sont remises en question \citep{Russel1994EarlyCrit, Barett2017Book, Siegel2018MetaAnalysisBig}. De là naît l'approche constructiviste :
celle-ci affirme que l'émotion est le produit d'un apprentissage en lien avec un contexte socioculturel qui prend appui sur deux dimensions fondamentales, la valence et l'excitation (\textit{arousal}), qui structurent les états internes. La valence spécifie si une expérience subjective est positive ou négative et l'activation définit le degré de stimulation physiologique. Ce paradigme est aussi qualifié de \textit{dimensionnel}, en opposition aux catégories discrètes d'Ekman.

En parallèle de ces deux perspectives sur l'émotion, une troisième a été développée~: l'approche évaluative.
Celle-ci énonce que ce ne sont pas les événements \textit{per se}, mais l'interprétation que chacun en fait, qui est à la source des émotions.
Selon les théories évaluatives, un stimulus -- un tremblement de terre, une silhouette dans le noir, une notification de rendez-vous manqué -- est évalué et produit une réponse adaptée, conscientisée comme la peur ou sous une autre forme.
Dans ce cadre, les émotions sont la représentation subjective d’un état synchronisé du corps et de la pensée, en relation forte avec une situation donnée. Cette vision de l'émotion n'est pas incompatible avec les deux précédentes. C'est moins tant par sa perspective théorique que par l'accent mis sur la perception et l'interprétation du contexte qu'elle se distingue des deux précédentes.

\subsection*{Tour d'horizon des corpus annotés}

Les émotions sont un processus qui se manifeste sous de nombreuses modalités
corporelles, cognitives et expressives.
D'abord étudiées à partir des expressions du visage, elles ont  par la suite été examinées dans l'expression langagière. Nous passons en revue quelques-uns des corpus annotés qui ont contribué à guider la réflexion théorique à ce sujet, en les mettant en relation avec les théories psychologiques que nous venons de décrire brièvement.
Le langage lui-même peut prendre différentes formes~; nous prenons le parti, en lien avec notre sujet, de ne considérer que les bases textuelles en écartant les corpus oraux et multimodaux.

En ouverture de cette section, nous avons présenté le modèle des émotions élémentaires d'\citet{Ekman1987UniversalsFace}, l'un des plus influents dans le domaine de l'étude des émotions. De fait, plusieurs corpus s'en inspirent au début des années 2000 pour définir leur schéma d'annotation, comme le relèvent \citet{Bostan2018CorpusReview} dans leur revue. À cet égard, nous pouvons citer les travaux d'\citet{Alm2005CorpusTales} et \citet{Strapparava2007SemEval07} qui reprennent directement les six catégories initialement avancées par Ekman.
Ce modèle de l'émotion reste par la suite une inspiration majeure, employé directement \cite{Li2017DailyDialog} ou avec quelques ajustements \cite{Mohammad2012CorpusTEC, Liu2019CorpusDens}\footnote{En particulier, certains corpus font appel à une variante du modèle d'Ekman développée par \citet{Plutchik1980EmotWheel1, Plutchik2001EmotWheel2}.}.
S'agissant de l'approche dimensionnelle, elle a également été appliquée aux données textuelles, quoique dans des proportions plus réduites \citep{Preotiuc2016CorpusFaceb, Buechel2017EmoBankDim}. Nous présentons dans le Tableau \ref{tab:corpus_review} les corpus les plus importants identifiés par \citet{Bostan2018CorpusReview} (taille > 10k exemples) auxquels nous en avons ajouté d'autres, francophones ou postérieurs à la revue.

\bgroup
\def\arraystretch{1.1}
\begin{table}[!ht]
    \centering
    \footnotesize
    \begin{tabular}{|L{2.7cm}|L{1.7cm} L{1.95cm} R{0.9cm} L{2.6cm}|}
        \hline
        Corpus & Granularité & \# Annotation & Taille & Source\\ \hline
        \textit{Corpus anglophones} & & & & \\
        Tales & phrases & 6$\,$C + V & 15~302 & \vspace{-8.5pt} \citet{Alm2005CorpusTales} \\
        CrowdFlower & tweets & 14$\,$C & 40~000 & \vspace{-8.5pt} \citet{VanPelt2012Crowdflower} \\
        TEC & tweets & 8$\,$C & 21~051 & \vspace{-8.5pt} \citet{Mohammad2012CorpusTEC} \\
        DailyDialog & conversation & 6$\,$C & 13~118 & \vspace{-8.5pt} \citet{Li2017DailyDialog} \\
        EmoBank & phrases & V + A + D & 10~548 & \vspace{-8.5pt} \citet{Buechel2017EmoBankDim} \\
        DENS & extraits litt. & 8$\,$C + N & 9~710 & \vspace{-8.5pt} \citet{Liu2019CorpusDens} \\
        GoEmotions & posts Reddit & 27$\,$C + N& 54~263 & \vspace{-8.5pt} \citet{Demszky2020GoEmot} \\
        XED & sous-titres & 8$\,$C + N + V & 23~940 & \vspace{-8.5pt} \citet{Ohman2020CorpusXED} \\
        x-enVENT & phrases & 15$\,$C + contexte & 1~001 & \vspace{-8.5pt} \citet{Troiano2022CorpusXenVENT} \\ \hline
        \textit{Corpus francophones} & & & & \\
        EMA & mots & 6~C + V + A & 1~286 & \vspace{-8.5pt} \citet{Gobin2017CorpusEMA} \\
        FANCat & mots & 10~C + V + A & 1~031 & \vspace{-8.5pt} \citet{Syssau2021CorpusFANCat} 
        \\
        TextToKids-Émotions & textes & 10~C + contexte & 1~594 & \vspace{-8.5pt} \citet{Etienne2024TextToKids} \\
        \hline
    \end{tabular}
    \caption{Sélection de ressources pour l'analyse des émotions. Les annotations font référence aux schémas suivants~: [$n\,$C] nombre de catégories discrètes~; [N] catégorie neutre~; [V] valence~; [A] excitation~; [D] dominance ; [contexte] annotation détaillée du contexte d'énonciation.}
    \label{tab:corpus_review}
\end{table}
\egroup

Ces dernières années, les travaux qui visent à identifier l'émotion tendent à employer des paradigmes d'annotation plus élaborés qu'auparavant. Annoté par contribution collective, le corpus CrowdFlower \citep{VanPelt2012Crowdflower} s'appuie sur 16 étiquettes non standards qui reprennent et élargissent les catégories d'Ekman. Le corpus GoEmotions, l'un des plus massifs à ce jour, adopte quant à lui un jeu de 28 étiquettes pour caractériser l'émotion dans des commentaires Reddit. Il fait suite aux travaux de \citeauthor{Cowen2021EmotSemSpace} (\citeyear{Cowen2021EmotSemSpace, Cowen2020EmotFacialNew1}; \citeauthor{Cowen2021EmotFacialNew2}, \citeyear{Cowen2021EmotFacialNew2}) qui, par le biais de nouveaux outils statistiques et notamment les réseaux de neurones, réimaginent la théorie des émotions de base de Paul Ekman \cite{Keltner2019EmotRevNow}. Privilégiant une approche différente, \citet{Troiano2022CorpusXenVENT} fondent le schéma d'annotation du corpus x-enVENT sur la théorie évaluative. En plus de 15 étiquettes pour représenter les états affectifs, ils en emploient 22 pour caractériser le processus d'évaluation du contexte émotionnel. Ces 22 dimensions ont chacune été évaluées sur une échelle de Likert de 1 à 5. La contrepartie d'une telle procédure est la taille du corpus, puisque seul un millier de phrases environ a pu être étiqueté.


Dans l'espace francophone, le volume de ressources textuelles annotées est sensiblement plus faible~: nous trouvons principalement des bases lexicales comme les corpus EMA \citep{Gobin2017CorpusEMA} et FANCat \citep{Syssau2021CorpusFANCat}. Un projet qui se distingue par son envergure est celui porté par \citet{Etienne2023TextToKidsThesis}, sous la direction de D. Battistelli et G. Lecorvé. Dans le cadre de cette thèse a été compilé un ensemble d'articles de presse pour enfants qui ont ensuite été annotés librement selon 10 catégories de l'émotion, mais également selon le type d'entités en jeu et selon une typologie développée par \citet{Blanc2017Psycholing1} et \citet{Micheli2025Psycholing2}.

Au regard de ce bref tour d'horizon, il nous semble clair que le champ d'étude de l'expression de l'émotion dans le texte est encore très ouvert. Tant d'un point de vue théorique --~nous avons commencé par montrer que le concept même d'émotion était sujet à débat, sans même mentionner les travaux qui traitent de l'aspect purement linguistique de la question~-- que du point de vue des ressources disponibles, limitées pour la langue française. C'est dans ce contexte que nous présentons ci-après un nouveau corpus annoté, basé sur une ressource textuelle et une procédure d'étiquetage originales.

\section{Données expérimentales}

\subsection*{Un corpus d'opinions face aux innovations}

Comme présenté en introduction, les données linguistiques que nous étudions ont été produites dans un environnement industriel. Celles-ci sont issues de la plate-forme communautaire Yoomaneo\footnote{\href{https://webapp.yoomaneo.com/}{https://webapp.yoomaneo.com/}}, développée en 2020 par Ixiade \citep{Masson2023Yoomaneo}. Il s’agit d’une application web qui vise à mesurer l'acceptabilité des concepts innovants en les soumettant au jugement de sa base d'utilisateurs.
Malgré une visée motivée par des considérations économiques, la variété des innovations testées justifie un intérêt plus large d'ordre scientifique.

En pratique, tout projet débute par un questionnaire d'éligibilité, qui permet de s’assurer que le profil des participants correspond à celui ciblé. Si l'innovation examinée est un nouveau tableau électrique, ce sont des électriciens qui sont recherchés. Les critères de sélection peuvent avoir trait au métier, à l'âge, à la catégorie socioprofessionnelle, etc.
Suite à cela, l'innovation testée est exposée via une courte vidéo et une série de questions ouvertes est posée.
Celles-ci suivent un protocole standard, réadapté en fonction des concepts et des interrogations spécifiques des clients porteurs des projets. Les questions explorent la perception spontanée, la compréhension du concept, de son usage et de ses fonctions, l’adéquation aux valeurs et aux structures sociales. Toutes demandent aux participants de rédiger un message -- un \textit{post} -- pour répondre.
Lorsque le projet est terminé les participants sont dédommagés par une carte cadeau dont le montant varie selon l'implication et la nature du projet.

Le corpus ainsi formé regroupe 2446 posts, correspondant à 22 concepts testés, produits par 238 membres de la plate-forme distincts. Pour l’annotation selon le protocole décrit ci-après, les posts sont divisés en verbatims\footnote{Segment de texte, généralement de la taille d’une phrase.}, formant un ensemble total de 4980 verbatims.
Le corpus fait actuellement l'objet d'une procédure d'anonymisation en vue de sa publication.






\subsection*{La composante évaluative recontextualisée}

Pour analyser l'émotion dans les productions textuelles, nous nous appuyons sur le modèle théorique \textsc{eminosa}. Celui-ci résulte des travaux de recherches réalisés par le laboratoire LIP/PC2S\footnote{Laboratoire Inter-universitaire de Psychologie. Personnalité, Cognition, Changement Social.} en partenariat avec Ixiade, menés successivement par \citet{Sbai2013Eminosa1}, \citet{Dupre2016Eminosa2}, et \citet{Loeser2019EMINOSA}. Dans ses fondements, il est inspiré du modèle multi-componentiel de l’émotion développé par \citet{Sherer2001Appraisal}\footnote{D'autres travaux ont également servi de référence pour \textsc{eminosa}, dont ceux de \citet{Frijda1986EmotDims} et \citet{Roseman2013CoherenceCoping}.}. Ce dernier divise l'émotion en cinq composantes qui interagissent entre elles : évaluative, conative, expressive, physiologique et subjective. Outre la composante évaluative et la composante subjective qui caractérise les états internes, les trois autres prennent en charge le mouvement du corps\footnote{La composante conative intègre la projection au mouvement --~l'action telle qu'elle est conçue en amont~; la composante motrice les mouvements involontaires (rythme cardiaque, sudation) et la composante expressive les mouvements volontaires (expression faciale, posture).}. Ce modèle donne un rôle prépondérant à la composante évaluative, aussi qualifiée de \textit{cognitive} en raison de ses liens avec d'autres mécanismes mentaux comme l'attention, la mémoire ou le raisonnement.

Dans le cadre de nos travaux, nous nous sommes nous-même intéressé plus particulièrement à la composante évaluative. Tout comme le modèle de \citeauthor{Sherer2001Appraisal}, \textsc{eminosa} subdivise cette composante --~que nous nommons \emin{}~-- en plusieurs dimensions~:
\begin{itemize}
    \item la Familiarité – le concept innovant testé évoque des choses connues et, par extension, son fonctionnement est bien compris~;
    \item l’Agréabilité – indépendamment de l’Utilité, le concept est plaisant / attractif\footnote{Cette définition inclut les propriétés ergonomiques et la facilité / difficulté d’utilisation.}~;
    \item l’Utilité – le concept remplit une fonction ou améliore l’existant ;
    \item la Légitimité – Le concept est en accord avec les valeurs personnelles / professionnelles.
\end{itemize}

Là où le modèle multi-componentiel et les autres modèles évaluatifs envisagent la réaction émotionnelle pour un stimulus indifférencié, \textsc{eminosa} a été développée spécifiquement pour caractériser les réactions face aux concepts innovants. C'est ainsi, par exemple, que la dimension <<~adéquation aux buts en cours~>> (\textit{goal relevance}), telle que proposée par \citeauthor{Sherer2001Appraisal}, prend la forme de l'Utilité. Par ailleurs, le modèle \textsc{eminosa} a été testé pour lui-même lors de plusieurs expériences indépendantes \citep{Loeser2019EMINOSA}.

\subsection*{Annotation de la composante évaluative dans le texte}

Nous décrivons ci-après le processus d'annotation des opinions selon les dimensions de la composante \emin{}. Celui-ci a fait l'objet d'une communication antérieure, dans laquelle il est développé plus en détails en même temps que des analyses préliminaires \citep{Noblet2025EminosaNoise}.
Comme pour d'autres corpus, la démarche d'annotation textuelle est motivée par la problématique de mieux expliquer et référencer les affects -- dans notre cas envers des concepts innovants.

Pour identifier la manifestation des dimensions \emin{} dans notre corpus, nous avons recouru à des annotateurs humains. Pour les quatre dimensions, les annotateurs ont attribué une valeur à chaque verbatim~:
\begin{itemize}
    \item +1 si la dimension est présente positivement (pour l’Utilité par exemple, l’évaluateur exprime que le concept est utile)~;
    \item -1 si la dimension est présente négativement (le concept est inutile)~;
    \item 0 si la dimension est absente ou présente de manière ambigüe.
\end{itemize}

Pour l’Exemple~\ref{ex:annotation}, un annotateur aura tendance à associer la valeur +1 à l’Agréabilité (intuitif, facile d’utilisation), -1 à l’Utilité (ne résout pas le problème du poids), et 0 pour la Familiarité et la Légitimité.
\begin{exe}
    \item \textit{Cela semble très intuitif, facile d’utilisation mais cela ne résout pas le problème du poids.}
    \label{ex:annotation}
\end{exe}

Quatre études pilotes et une campagne d'annotation à large échelle ont été menées. Pour cette dernière, 6 annotateurs ont été recrutés qui ont tous traité l'ensemble des verbatims de 21 des 22 projets\footnote{Le projet écarté est l'un des premiers ayant été déployé sur la plate-forme, selon des modalités notablement différentes de ceux qui ont suivi.}. La plupart des corpus annotés s'appuient sur un nombre réduit d'annotateurs~; en mobiliser six permet de mieux appréhender la variabilité des pratiques d'annotation. En incluant les études pilotes, 4873 verbatims ont été annotés donnant lieu à 32~732 annotations.

\section{Analyses et résultats}
\label{sec:results}

\subsection*{Statistiques générales}

\begin{figure}[ht]
        \hfill%
        \begin{subfigure}[b]{0.45\linewidth}
            \centering
            \includegraphics[scale=0.65]
            {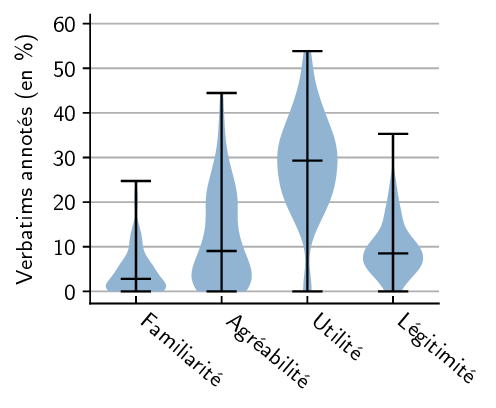} 
            \caption{Verbatims annotés positivement} 
            \label{fig:violin_pos} 
            \vspace{4ex}
        \end{subfigure}
        \hfill%
        \begin{subfigure}[b]{0.45\linewidth}
            \centering
            \includegraphics[scale=0.65]
            {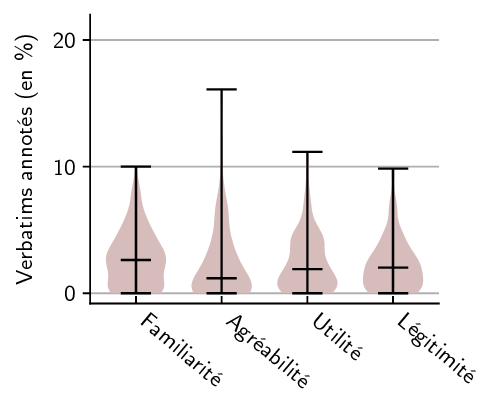} 
            \caption{Verbatims annotés négativement} 
            \label{fig:violin_neg} 
            \vspace{4ex}
        \end{subfigure}
        \hfill%
    \vspace{-0.5cm}
    \caption{Proportion des verbatims annotés, par annotateur et par projet.}
    \label{fig:violins}
\end{figure}

Préalablement à des analyses plus fines, nous présentons quelques traitements statistiques qui donnent un aperçu général du corpus annoté.
En première observation, nous constatons que les annotations \emin{} sont fortement déséquilibrées. Comme le montre la Figure~\ref{fig:violins}, la part de verbatims annotés positivement et négativement varie grandement en fonction des dimensions, projets et annotateurs considérés. De plus, les annotations négatives sont très peu fréquentes~: seuls 3\% des verbatims environ sont annotés négativement, toutes dimensions confondues, comme l'indique la Figure~\ref{fig:violin_neg}. Remarquons également que la dimension de Familiarité dans son ensemble est faiblement mise en évidence. Inversement, l'Utilité domine les annotations, se manifestant deux à trois fois plus que les autres dimensions. Cette dernière tendance semble appuyer le modèle évaluatif récemment développé par \citet{Moors2017GoalAppraisDim}~: cette chercheuse soutient que l'utilité attendue des actions disponibles occupe une place centrale dans la réaction émotionnelle.

Une autre spécificité de notre corpus est la grande part d'exemples
qui ne sont pas annotés
-- qu'aucun annotateur n'a étiquetés positivement ou négativement, pour une catégorie donnée -- s'élevant à 45\% des verbatims pour l'Utilité et jusqu'à 76\% pour la Familiarité.
Les autres corpus que nous avons référencés appliquent généralement une étape de sélection des exemples pertinents pour l'analyse des émotions \citep{Demszky2020GoEmot, Mohammad2012CorpusTEC, Etienne2024TextToKids}. Nous avons préféré maintenir notre corpus en l'état~: identifier quels verbatims doivent ou non être annotés est une partie intégrante de la problématique de la reconnaissance des émotions dans le texte. Nous privilégions ainsi le critère écologique -- des données pleinement représentatives de leur contexte d'énonciation -- à celui du contraste -- des dimensions bien démarquées.

En outre, nous avons relevé un fort désaccord entre les annotations. Le degré d'accord pour chaque dimension mesuré à l'aide de l'alpha de Krippendorff \citep{Krippendorff2011Comput} est représenté dans la première ligne du Tableau~\ref{tab:accord}. Notons toutefois que ce désaccord n'est pas total. Si nous ignorons les annotations nulles et que nous jugeons seulement l'accord entre annotations positives et négatives (deuxième ligne du Tableau ~\ref{tab:accord}), l'accord est très bon ($\alpha \geq 0,8$). À l'inverse, si nous ignorons la polarité et que nous tenons uniquement compte de la présence ou de l'absence des dimensions pour un verbatim donné (troisième ligne du Tableau ~\ref{tab:accord}), nous observons un fort désaccord, reflétant le score global.

\bgroup
\def\arraystretch{1.1}
\begin{table}[!ht]
    \centering
    \begin{tabular}{|L{1.8cm}|C{1.8cm} C{1.8cm} C{1.8cm} C{1.8cm}|}
        \hline
        Modalité & Familiarité & Agréabilité & Utilité & Légitimité \\
        \hline
        Global & $0,29$ & 0,31 & 0,53 & 0,38 \\
        Polarité & 0,80 & 0,93 & 0,91 & 0,92 \\
        Pertinence & 0,30 & 0,28 & 0,51 & 0,36 \\
        \hline
    \end{tabular}
    \caption{Accord inter-annotateurs pour les dimensions d’\emin{}, pour l’ensemble des annotations, suivant différentes modalités de mesure.}
    \label{tab:accord}
\end{table}
\egroup

Après avoir examiné les données en surface, lors d'une communication précédente \citep{Noblet2025EminosaNoise}, nous avons émis l’hypothèse qu’il existe un modèle probabiliste sous-jacent qui gouverne les pratiques d’annotation~; les annotations \emin{}, bien qu’aléatoires, suivraient des tendances statistiques explicables. Des traits linguistiques, comme un lexique explicitement associé aux dimensions\footnote{En se référant à la typologie d’\citet{Etienne2020AnnotChildStory}, l’appellation dimensions \textit{désignées} pourrait être employée .} ou un fort degré de confiance apparent semble aller de pair avec une annotation unanime. Schématiquement, plus une dimension est marquée avec intensité dans un verbatim donné, plus il sera annoté. Le Tableau~\ref{tab:accord_exemple} reproduit des exemples que nous avions donné à l'appui de cette hypothèse, annoté à différentes proportions pour la dimension de Légitimité.

\bgroup
\def\arraystretch{1.1}
\begin{table}[ht]
    \centering
    \begin{tabular}{|C{2cm}  L{8cm}|}
        \hline
         Légitimité & {\small{Exemples}} \\
         \hline
         1,00 (6/6) & \textit{Cette solution s'appuie sur une approche "objet connecté", clairement dans l'air du temps.} \\
         0,83 (5/6) &  \textit{De plus, si ce produit peut se décliner pour corriger nos postures au quotidien même au travail, cela répond au souci de préservation de sa santé sur le long-termes.} \\
         0,67 (4/6) &  \textit{On est en plein dans les objects connectés du futurs, reliés complètement à nos smartphones (ce n'est pas un mal)} \\
         0,50 (3/6) & \textit{Iot, bien être, sport à la maison} \\
         0,33 (2/6) &  \textit{Préventif : permet de soigner mais aussi d'anticiper les douleurs}  \\
         0,17 (1/6) &  \textit{Repond a un besoin, propose une solution complete, suscite l'interet.} \\
         \hline
    \end{tabular}
    \caption{Exemples correspondant à différents niveaux de Légitimité moyenne positive}
    \label{tab:accord_exemple}
\end{table}
\egroup

\subsection*{Modélisation des annotations par réseaux Transformers}

Pour éprouver l'hypothèse selon laquelle les dimensions \emin{} sont annotées en suivant un gradient stable, nous avons affiné, sur la base des annotations moyennées, plusieurs modèles de type Transformer pré-entraînés \citep{Vaswani2017Transform}. Si ces modèles sont en mesure de retrouver les proportions annotées sur des corpus test, cela attesterait de la présence d'indices linguistiques qui motiveraient la variation dans les jugements.

À cette fin, nous nous sommes basés sur les modèles Flaubert (flaubert-base-cased, 138~M de paramètres) \citep{Le2020ModelFlaubert} et GTE (gte-multilingual-base, 305~M de paramètres) \citep{Zhang2024ModelAlibaba}, développé par la société Alibaba. Ce dernier a été choisi en raison de ses bonnes performances au test d'évaluation MTEB \cite{Ciancone2024MTEBEval}. À partir de ces modèles, nous avons conçu trois architectures :
\begin{enumerate}[(1)]
    \item modèle Flaubert gelé connecté à un classifieur logistique~;
    \item modèle GTE gelé connecté à un classifieur logistique~;
    \item modèle GTE gelé à l'exception de la dernière couche, connecté à un classifieur logistique.
\end{enumerate}

\begin{figure}[p] 
        \begin{subfigure}[b]{0.48\linewidth}
            \centering
            \includegraphics[scale=0.7]
            {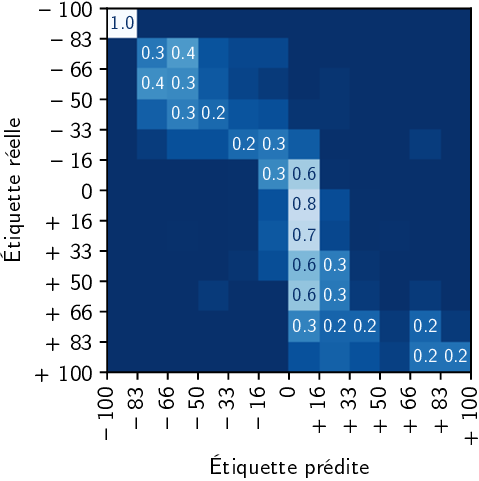} 
            \caption{Familiarité} 
            \label{fig:confusion_ND_tuned_alibaba_F} 
            \vspace{4ex}
        \end{subfigure}
        \hfill%
        \begin{subfigure}[b]{0.48\linewidth}
            \centering
            \includegraphics[scale=0.7]
            {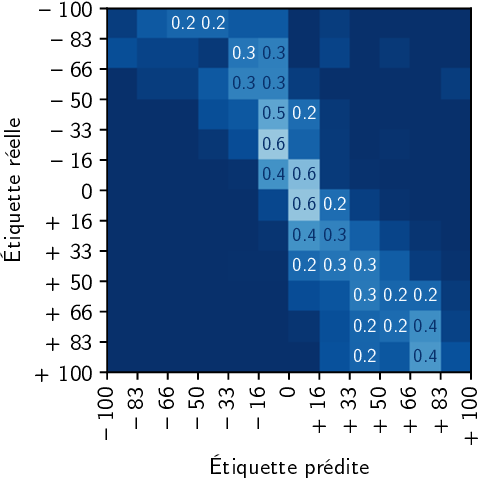} 
            \caption{Agréabilité} 
            \label{fig:confusion_ND_tuned_alibaba_A} 
            \vspace{4ex}
        \end{subfigure}
        \begin{subfigure}[b]{0.48\linewidth}
            \centering
            \includegraphics[scale=0.7]
            {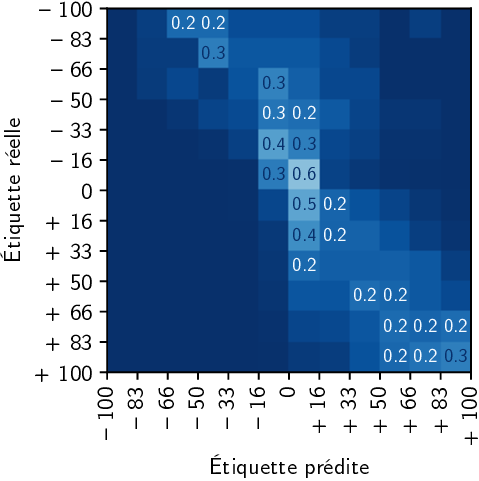} 
            \caption{Utilité} 
            \label{fig:confusion_ND_tuned_alibaba_U} 
        \end{subfigure}
        \hfill%
        \begin{subfigure}[b]{0.48\linewidth}
            \centering
            \includegraphics[scale=0.7]
            {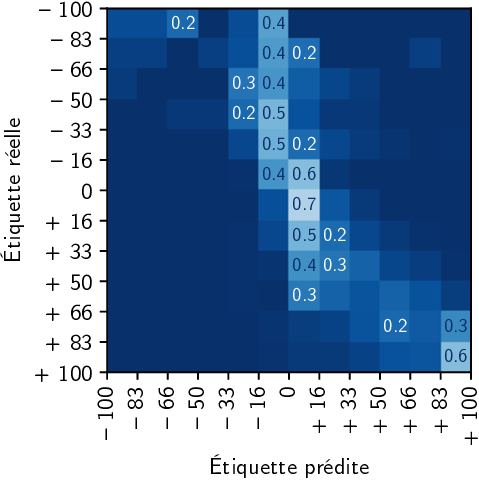} 
            \caption{Légitimité} 
            \label{fig:confusion_ND_tuned_alibaba_L} 
        \end{subfigure}
        \par\bigskip
        \begin{subfigure}{\linewidth}
            \centering
            \includegraphics[scale=0.7]
            {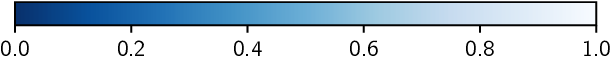}
        \end{subfigure}
        \par\medskip
    \caption{Modèle GTE -- dernière couche entraînée}
    \label{fig:confusion_ND_tuned_alibaba}
\end{figure}

Les modèles ont été entraînés séparément sur chaque dimension en suivant un protocole de validation croisée à l'échelle des projets -- 4 projets ont servi successivement de tests tandis que les autres ont contribué à l'entraînement.
Nous représentons les performances d'annotation automatique sous la forme de matrices de confusion --  la configuration (3) est illusrée en Figures
\ref{fig:confusion_ND_tuned_alibaba}, les configurations (1) et (2) sont données en annexe (Figures~\ref{fig:confusion_ND_frozen_flaubert}, \ref{fig:confusion_ND_frozen_alibaba}).

Remarquons de prime abord que le modèle Flaubert seul n'est aucunement en mesure de distinguer les différentes valeurs d'annotation, quelle que soit la dimension considérée\footnote{Comme le modèle GTE, il est toutefois possible de l'affiner pour trouver des performances correctes.}. À l'inverse, le modèle GTE, même sans entraînement supplémentaire, est en mesure de produire des représentations qui permettent déjà de discriminer grossièrement des gradients d'annotation. Nous pouvons en effet constater que les annotations produites suivent la diagonale des matrices de confusion. À l'exception de la Familiarité, le modèle a néanmoins tendance à beaucoup délaisser les étiquettes négatives (cadrans supérieurs des matrices) au profit de la classe majoritaire, l'annotation nulle.

Les performances les plus élevées sont obtenues en donnant un <<~coup de pouce~>> au modèle GTE, en entraînant sa dernière couche. Dans cette configuration, nous constatons que les étiquettes produites par le modèle suivent plutôt fidèlement la diagonale, ce qui indique que le modèle parvient à retrouver les proportions d'annotation (voir Figure~\ref{fig:confusion_ND_tuned_alibaba}). En particulier, pour des verbatims annotés positivement en forte proportion (extrémité inférieure droite), l'annotation nulle est correctement abandonnée par le modèle au profit d'une annotation marquée positivement (pour l'utilité, les prédictions du modèle sont supérieures à 0,33 en majorité, voir Figure~\ref{fig:confusion_ND_tuned_alibaba_U}).

Le Tableau~\ref{tab:llm_scores} présente une estimation des performances du modèle GTE affiné (confirguration 3), en délimitant les valeurs par des seuils arbitraires. Bien que cette estimation ne permette pas de rendre compte avec justesse des résultats obtenus car nous présupposons une échelle continue, elle offre un aperçu synthétique plus directement interprétable. Comme nous l'avons précédemment remarqué, à l'exception de la Familiarité, toutes les autres dimensions sont faiblement annotées par le modèle automatique pour la polarité négative~; la précision est bonne mais le rappel très faible (inférieur à 0,25). La dimension de Familiarité se distingue des trois autres~: elle est mieux annotée pour la polarité négative que pour la polarité positive. Cette tendance s'explique par des verbatims annotés en plus grand nombre pour la polarité négative. Nous retrouvons ici la spécificité du corpus, constitué d'opinions au sujet de concepts innovants souvent peu familiers. En dernier lieu, notons que la Légitimité est la dimension la plus problématique -- malgré une bonne précision, le rappel est faible pour les étiquettes positives (49,7) et très faible pour les étiquettes négatives (10,5). Ce rappel peu élevé illustre là encore la propension du modèle à adopter l'étiquette nulle majoritaire dans le corpus, visible à travers la bande verticale en Figure~\ref{fig:confusion_ND_tuned_alibaba_L}.

Ces résultats, bien qu'encourageants, sont encore trop approximatifs pour constituer une conclusion définitive à nos recherches. Nous avons entrepris des réflexions pour quantifier avec justesse les sorties des systèmes neuronaux. Parallèlement, un autre axe de travail porte sur l'exploration approfondie des critères influençant les décisions des modèles.

\bgroup
\def\arraystretch{1.1}
\begin{table}[!ht]
    \centering
    \begin{tabular}{|l|C{1.8cm} C{1.8cm} C{1.8cm} C{1.8cm}|}
         \hline
         Métrique &  Familiarité  &  Agréabilité & Utilité & Légitimité \\
         \hline
         Précision globale     &  $95,2$  &  $85.1$  &  $80.0$ &  $89.1$\\
         Précision \texttt{+1} &  $71,4$  &  $78.6$  &  $82.3$ &  $77.4$\\
         Rappel    \texttt{+1} &  $21,6$  &  $55.2$  &  $71.8$ &  $49.7$\\
         Précision \texttt{-1} &  $80,5$  &  $80.9$  &  $76.2$ &  $48.8$\\
         Rappel    \texttt{-1} &  $69,8$  &  $24.0$  &  $20.3$ &  $10.5$\\
         Précision \texttt{0}  &  $95,9$  &  $86.2$  &  $78.7$ &  $90.4$\\
         Rappel    \texttt{0}  &  $99,1$  &  $95.7$  &  $89.2$ &  $97.6$\\
         \hline
    \end{tabular}
    \caption{Modèle GTE affiné -- Scores de précision et rappel, pour chaque dimension et valeur d'\emin{}, avec des seuils à -1/3 et +1/3 pour segmenter les valeurs.}
    \label{tab:llm_scores}
\end{table}
\egroup

\section{Conclusion}

Cet article propose plusieurs contributions dans le champ de l’annotation et de la modélisation des émotions. En premier lieu, le jeu d'étiquettes sur lequel nous nous appuyons est fondé sur une conception de l'émotion aujourd'hui peu exploitée, l'approche évaluative. Cette approche, qui complémente les conceptions traditionnelles des émotions, permet de mieux prendre en compte la relation au stimulus -- dans notre cas un concept innovant.
De plus, en adoptant une démarche sans pré-filtrage et en recourant à six annotateurs pour traiter les mêmes unités, nous accédons à une large variabilité dans les jugements. Nous émettons l'hypothèse que cette variation observée est le reflet d'un gradient d'intensité dans l'expression de l'émotion~: plus une dimension de l'émotion se manifeste clairement, plus elle aura tendance à être annotée et inversement. Pour tester cette hypothèse, nous entreprenons de modéliser le processus d'annotation à l'aide de modèles de langues affinés. Ces derniers montrent des résultats probants, indiquant à la fois que la variabilité des annotations est gouvernée par des tendances stables et que les modèles de langue sont en mesure de porter une conception évaluative des émotions.

Notre corpus présente néanmoins plusieurs limitations.
Il repose exclusivement sur des discours écrits en réaction à un stimulus distant -- une vidéo de concept innovant.
Ce mode d'interaction peut constituer une barrière à l'engagement émotionnel, notamment lorsque les enjeux perçus sont faibles.
En outre, bien que les objets physiques et les services représentent une part significative de notre environnement, d’autres sources potentielles d’émotion restent exclues. En particulier, ne sont pas pris en compte les rapports interpersonnels, qui constituent un aspect fondamental de l’émotion.

Nos travaux futurs viseront à approfondir la validation de notre hypothèse. L’opacité des réseaux de neurones constitue un enjeu majeur, rendant nécessaire le développement de méthodes algorithmiques permettant d’analyser les choix effectués par ces modèles et d’identifier les structures linguistiques sous-jacentes. En complément, nous envisageons un retour aux méthodologies linguistiques afin d’affiner l’analyse, en apportant une précision explicative que les outils statistiques ne permettent pas d’atteindre.

\bibliographystyle{coria-taln2025}
\bibliography{references}


\appendix
\section*{Annexes}

Nous présentons ici des figures complémentaires illustrant les performances des modèles Flaubert gelé et GTE gelé (voir Section \ref{sec:results}).

\begin{figure}[p] 
        \begin{subfigure}[b]{0.48\linewidth}
            \centering
            \includegraphics[scale=0.7]
            {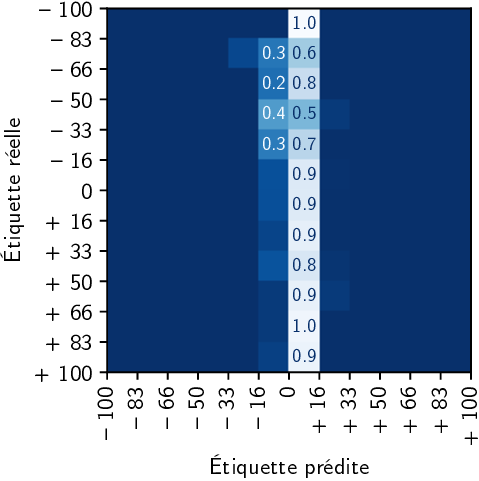} 
            \caption{Familiarité} 
            \label{fig:confusion_ND_frozen_flaubert_F} 
            \vspace{4ex}
        \end{subfigure}
        \hfill%
        \begin{subfigure}[b]{0.48\linewidth}
            \centering
            \includegraphics[scale=0.7]
            {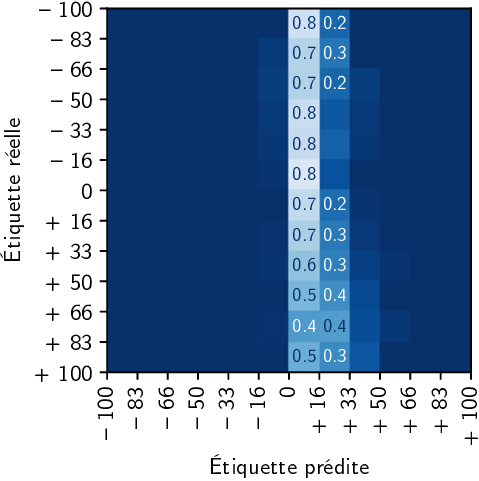} 
            \caption{Agréabilité} 
            \label{fig:confusion_ND_frozen_flaubert_A} 
            \vspace{4ex}
        \end{subfigure}
        \begin{subfigure}[b]{0.48\linewidth}
            \centering
            \includegraphics[scale=0.7]
            {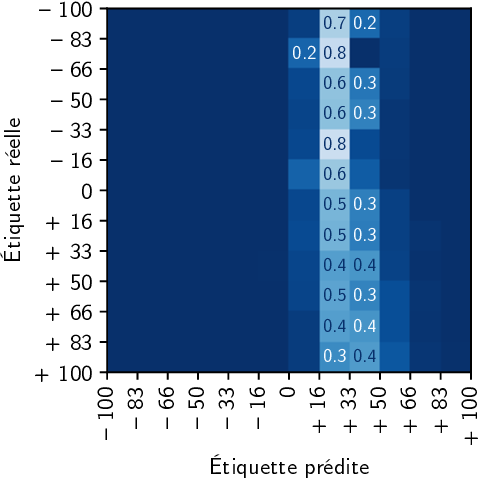} 
            \caption{Utilité} 
            \label{fig:confusion_ND_frozen_flaubert_U} 
        \end{subfigure}
        \hfill%
        \begin{subfigure}[b]{0.48\linewidth}
            \centering
            \includegraphics[scale=0.7]
            {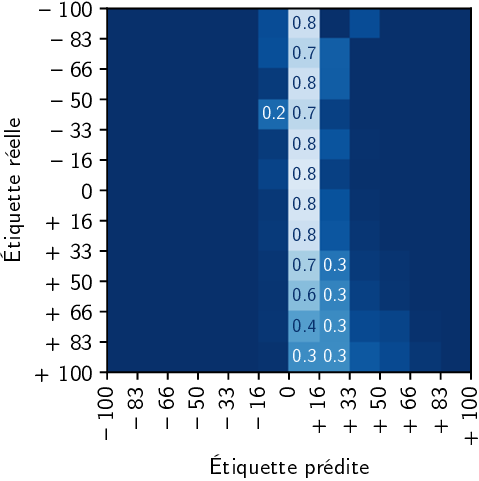} 
            \caption{Légitimité} 
            \label{fig:confusion_ND_frozen_flaubert_L} 
        \end{subfigure}
        \par\bigskip
        \begin{subfigure}{\linewidth}
            \centering
            \includegraphics[scale=0.7]
            {FIGURES/colorbar_h.eps}
        \end{subfigure}
        \par\medskip
    \caption{Modèle Flaubert -- modèle gelé}
    \label{fig:confusion_ND_frozen_flaubert}
\end{figure}

\begin{figure}[p] 
        \begin{subfigure}[b]{0.48\linewidth}
            \centering
            \includegraphics[scale=0.7]
            {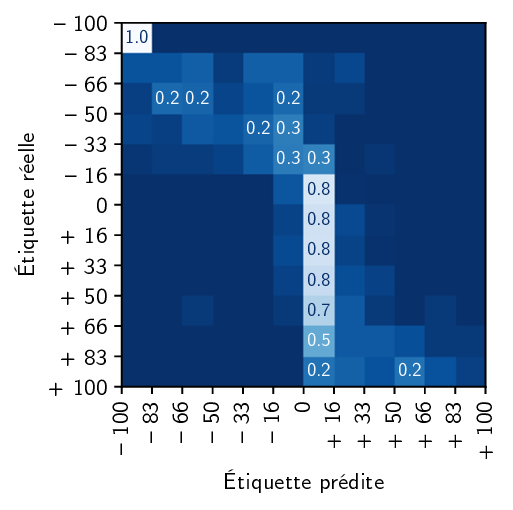} 
            \caption{Familiarité} 
            \label{fig:confusion_ND_frozen_alibaba_F} 
            \vspace{4ex}
        \end{subfigure}
        \hfill%
        \begin{subfigure}[b]{0.48\linewidth}
            \centering
            \includegraphics[scale=0.7]
            {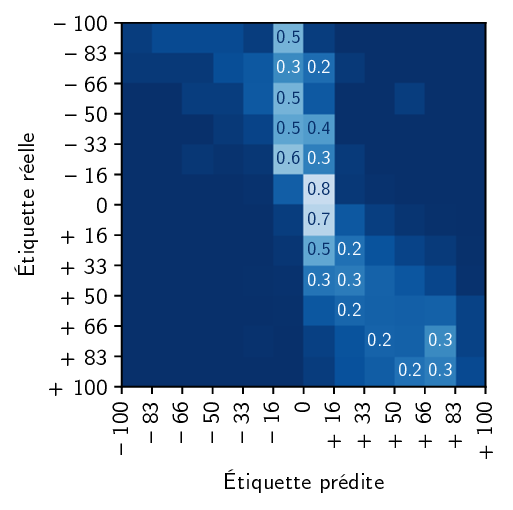} 
            \caption{Agréabilité} 
            \label{fig:confusion_ND_frozen_alibaba_A} 
            \vspace{4ex}
        \end{subfigure}
        \begin{subfigure}[b]{0.48\linewidth}
            \centering
            \includegraphics[scale=0.7]
            {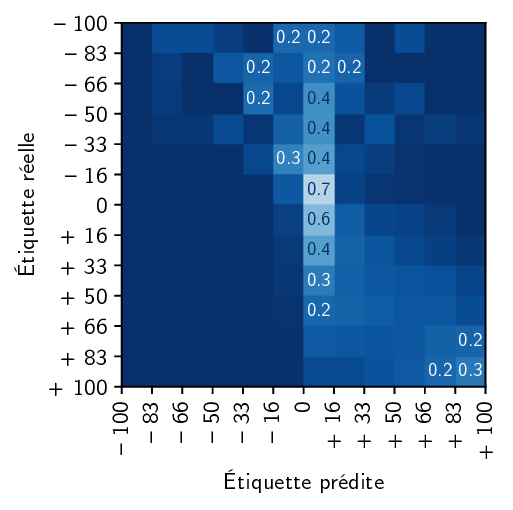} 
            \caption{Utilité} 
            \label{fig:confusion_ND_frozen_alibaba_U} 
        \end{subfigure}
        \hfill%
        \begin{subfigure}[b]{0.48\linewidth}
            \centering
            \includegraphics[scale=0.7]
            {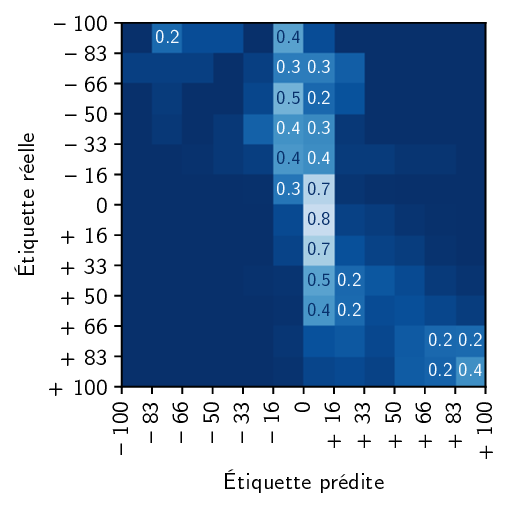} 
            \caption{Légitimité} 
            \label{fig:confusion_ND_frozen_alibaba_L} 
        \end{subfigure}
        \par\bigskip
        \begin{subfigure}{\linewidth}
            \centering
            \includegraphics[scale=0.7]
            {FIGURES/colorbar_h.eps}
        \end{subfigure}
        \par\medskip
    \caption{Modèle GTE -- modèle gelé}
    \label{fig:confusion_ND_frozen_alibaba}
\end{figure}

\end{document}